\documentclass[a4paper,fleqn]{cas-dc}

\usepackage{booktabs,tabularx,xcolor,array,rotating,pifont,textcomp,listings}
\usepackage{amsmath,amssymb,amsfonts}
\usepackage[numbers,sort&compress]{natbib}

\def\tsc#1{\csdef{#1}{\textsc{\lowercase{#1}}\xspace}}
\tsc{WGM}
\tsc{QE}
\tsc{EP}
\tsc{PMS}
\tsc{BEC}

\newcolumntype{Y}{>{\raggedright\arraybackslash}X}

\begin{document}
\let\WriteBookmarks\relax
\def\floatpagepagefraction{.8}
\def\textpagefraction{.2}

\shorttitle{Digital Twin and Agentic AI for Wild Fire Disaster Management: Intelligent Virtual Situation Room}


\title [mode = title]{Digital Twin and Agentic AI for Wild Fire Disaster Management: Intelligent Virtual Situation Room}

\author[1]{Mohammad Morsali}

\ead{mohammadmorsali1381@gmail.com} 

\author[2]{Siavash H.Khajavi}
\cormark[2] 
\ead{siavash.khajavi@aalto.fi}

\affiliation[1]{organization={Independent Researcher}}

\affiliation[2]{organization={Department of Industerial Engineering and Management, Aalto University},  
   city={Espoo},
   postcode={02150},
   country={Finland}}


\begin{abstract}
According to the United Nations, wildfire frequency and intensity are projected to increase by approximately 14\% by 2030 and 30\% by 2050 due to global warming, posing critical threats to life, infrastructure, and ecosystems. Conventional disaster-management frameworks rely on static simulations and passive data acquisition, which hinders their ability to adapt to arbitrarily evolving wildfire episodes in real-time. To address these limitations, we introduce the Intelligent Virtual Situation Room (IVSR), a bidirectional Digital Twin(DT) platform augmented by autonomous AI agents.  The IVSR continuously ingests multisource sensor imagery, weather data, and 3D forest models to create a live virtual replica of the fire environment. A similarity engine powered by AI aligns emerging conditions with a precomputed Disaster Simulation Library, bringing down and calibrating intervention tactics under the watchful eyes of experts. Authorized action—ranging from UAV redeployment to crew reallocation—is cycled back through standardized procedures to the physical layer, completing the loop between response and analysis. We validate IVSR through detailed case-study simulations provided by an industrial partner, demonstrating capabilities in localized incident detection, privacy-preserving playback, collider-based fire‐spread projection, and site-specific ML retraining. Our results indicate marked reductions in detection-to-intervention latency and more effective resource coordination versus traditional systems. By uniting real-time bidirectional DTs with agentic AI, IVSR offers a scalable, semi-automated decision-support paradigm for proactive, adaptive wildfire disaster management.

\end{abstract}

\begin{keywords}
Digital Twin \sep Wildfire Management \sep Disaster Response \sep AI Agents \sep Smart Cities
\end{keywords}

\maketitle 


\section{Introduction}

Natural disasters can devastate both the environment and urban infrastructure. Among natural disasters, wildfires pose a particularly serious threat due to the extensive damage they cause. In recent years, the frequency and severity of wildfires have increased due to climate change \cite{Wasserman2023}. Wildfires have led to billions in economic losses, irreversible ecological destruction, and the loss of lives and livelihoods. The 2023 Maui wildfires in Hawaii, for instance, resulted in over 100 deaths and destroyed nearly the entire town of Lahaina \cite{Kormann2023Maui}. In 2024, California experienced 8,024 wildfires, burning a total of 1,050,012 acres, destroying 1,716 structures, and resulting in one fatality \cite{calfire2024incidents}. Economic losses from the 2024 California wildfires are estimated between \$250 billion and \$275 billion, making it one of the costliest wildfire disasters in U.S. history \cite{danielle2025accuweather}. In Europe, the 2021 wildfires in Greece and Turkey burned hundreds of thousands of hectares, straining national emergency services and displacing communities \cite{aljazeera2021wildfires,nasa2021fire}. In Greece, approximately 50,000 hectares were burned on Evia Island   \cite{nasa2021fire}. In Turkey, over 95,000 hectares were affected, marking the worst wildfires in at least a decade \cite{aljazeera2021wildfires}.

 Despite advances in detection and modeling, the “golden window” of response—the critical early hours when intervention is most effective—is often missed. One of the primary reasons for this is the fragmented and inefficient data flow between various actors involved in wildfire management. Delays in situational awareness, limited interoperability between systems, and poor visibility into ground conditions hinder the ability of decision-makers to coordinate and deploy resources effectively. The result is a slow and reactive response, rather than a proactive, adaptive approach that could prevent escalation.

Various novel technologies may hold the key to implementing effective solutions for better collecting, analyzing, and responding to fast-moving catastrophic fire events. A digital twin (DT) is a real-time virtual representation of a physical entity or system, continuously updated with live data to reflect its current state \cite{abdul,Gri}. In early versions of the DT, it was used to improve system understanding and predict failures by utilizing a live digital model that collected data from the physical system with sensors \cite{Tao2018,KRITZINGER20181016}. During the past two decades, the DT concept expanded from aerospace and manufacturing to many other domains \cite{LU2020101837} – including civil infrastructure, energy grids, healthcare, and smart cities – as advances in IoT and computing made real-time data integration feasible on larger scales \cite{8477101}.

Smart city is used to describe an urban space enriched by ICT infrastructure and data-driven services. Modern digital cities utilize universal data capture and intelligent analysis, generating dynamic virtual simulations of urban areas (sometimes called city DTs) that mirror real-world status in real-time \cite{sss}. This may lay the groundwork for using digital cities for routine urban management and enhancing disaster resilience.

Artificial intelligence (AI) is a rapidly advancing field with wide-ranging implications across nearly all disciplines, including fire engineering \cite{Aras,Teh}. One emerging area within AI is the development of autonomous AI agents, where they are equipped with the ability to utilize tools and make decisions to complete tasks \cite{PICCIALLI2025128404}. These AI agents have the potential to become a critical resource in modern disaster relief, offering enhanced speed, intelligence, and coordination to support more informed decision-making by humans \cite{ALBAHRI2024109409}.

Yet, some fundamental questions remain unresolved, even as DT and AI technologies continue to evolve rapidly and become increasingly integrated into smart systems. This research investigates the following questions:

RQ1: What are the current use cases of bidirectional digital twins (BDTs) in the context of wildfire management?
RQ2: In what ways and configurations can an AI agent–based BDT assist with wildfire response?
RQ3: Based on case studies from industry and literature, what are the implications of an AI agent–based BDT for wildfire management?

To answer these questions, we review the literature, collect industry case deployments and finally conceptualize an intelligence virtual situation room (IVSR) enabled by a BDT for real-time wildfire management. IVSR is envisioned to integrate sensors, state-of-the-art AI-driven fire prediction models, and 3D representative models to establish a dynamic feedback loop between the physical environment and its digital counterpart. The IVSR concept improves early warning, better-informed decision-making, and proactive fire suppression in critical emergency scenarios by allowing for instant communication of actionable knowledge.

\section{Related Works}
\label{sec:relatedworks}
This section reviews the literature on digital cities and DT technologies, with a focus on their roles in wildfire disaster management.

The use of DTs in disaster management, especially in wildfires, is an emerging field. Previous studies have focused on flood prediction, earthquake simulation, and wildfire tracking using GIS-based mapping, hydrological models, and remote sensing. However, many existing solutions lack real-time data integration and interactive scenario simulations. Recent advancements in AI-driven anomaly detection, IoT-based monitoring, and climate modeling provide an opportunity to improve disaster preparedness and response \cite{sss}.

\subsection{Digital Cities}
\label{sec:digitalcities}

\subsubsection*{Digital City Infrastructure and Technologies}
Modern digital cities are built using advanced technology that works together \cite{MPD}. Sensor networks (environmental sensors, cameras, smart meters) and IoT devices continuously monitor urban conditions \cite{krishna}, from weather to traffic to structural health. Wireless communication systems also carry the data to improve the latency performance and provide real-time, citywide, and situational information \cite{app14209243}.

 The collected data is input into cloud-based centralized platforms. Geographic Information Systems (GIS) and 3D city modeling \cite{Marcooo} software play a key role by correlating data streams to spatial locations and creating interactive maps or virtual city models that reflect current conditions \cite{xu2024leveraginggenerativeaiurban,doi:10.1061/JUPDDM.UPENG-4650}. Also, ML/DL techniques are employed to detect(safety monitoring \cite{suan16219482,court2024usedigitaltwinssupport}) and predict outcomes \cite{MASCHLER2021127,alex,Wangggg}. Some examples include modeling the spread of fire \cite{Shadrin2024,fire7120482} or predicting traffic \cite{car}, anomaly detection \cite{https://doi.org/10.1049/2024/8821891}, and detecting abnormal trends\cite{ElShafeiy2023}, like an increase in water usage that may reveal a leakage \cite{10559837}. Also, more enabling technologies have been added, such as building DTs, which gives 3D representations of buildings \cite{10130166}; encryption and cyber security systems, which securely communicate data \cite{Xu2024,10.1145/3527049.3527140}; and edge computing, which processes the data locally \cite{Lawal,pan2018edgechainedgeiotframeworkprototype}.

These four key components—IoT sensing, communications networks, cloud/GIS, and AI-powered analytics—form the foundation of a digital city that can support services such as disaster management.

\subsubsection*{Role of Digital Cities in Disaster Management}

Digital city technologies have become critical tools in disaster management, as they improve how cities prepare for, respond to, and recover from emergencies by enabling more transparent communication and stronger coordination \cite{Ford,Wolf2022}. In this part, the main contributions made by digital cities toward disaster management will be introduced as follows:

Smart cities employ distributed sensors and analytics to detect threats or abnormal events immediately \cite{Adeel_2018}. For instance, the city's digital platform can sense a sudden drop in water pressure or surge in seismic movements and trigger instant warnings. A good example involved using a DT for the city to process incoming sensor data intelligently and trigger automated SMS warnings for emergencies (e.g., fires within a building or flash flooding \cite{Yuan_2022}).

Also, with the fusion of real-time multi-source data, digital cities generate a common operating picture during a crisis \cite{792de4e9f0b744e5b121ae278d3328a1,Adr}. City dashboards synthesize the feeds from CCTV cameras, traffic sensors, weather stations, and crowd reports to give responders live maps of dynamic circumstances. For example, analysts employed a DT of a city to process live surveillance cam feeds, identify congestion levels along evacuation routes, and route emergency vehicles through the most unobstructed courses \cite{buildings13051143}.

 So, The detailed digital model presentations help managers use "what-if" simulations for planning and disaster preparedness \cite{dogan2021digitaltwinbaseddisaster}. Scenarios such as a category-5 hurricane strike \cite{Accarino_2023} or chemical spill \cite{vondanwitz2024contaminantdispersionsimulationdigital} can be simulated and tested via digital models. Studies have shown that a high-fidelity DT of a city can simulate disaster scenarios – e.g., the spread of flooding and the damage resulting from a hypothetical flood –. By examining various scenarios (e.g., varying earthquake magnitudes \cite{inproceedings,8776632} and flood levels), authorities can refine emergency plans, such as adjusting evacuation zones and resource staging areas based on the simulation results \cite{792de4e9f0b744e5b121ae278d3328a1}. 

Resource coordination is one of the most widely used applications of digital cities. A digital city platform can monitor and coordinate vital resources within the urban network during a crisis \cite{jsan12030041}. The status of hospitals (availability and ER status), emergency shelters (degree of occupancy), utilities (power grid or water supply outages), fleets of response vehicles, and number and distribution of available first responders can be shown in real-time through integrated dashboards \cite{han2024designing,Palmieri}. The digital city system can facilitate the deployment of teams and equipment where necessary and adjust to crisis changes \cite{abraham2024evacuationmanagementframeworksmart}.

\begin{figure}[ht]
    \centering
    \includegraphics[width=0.8\columnwidth]{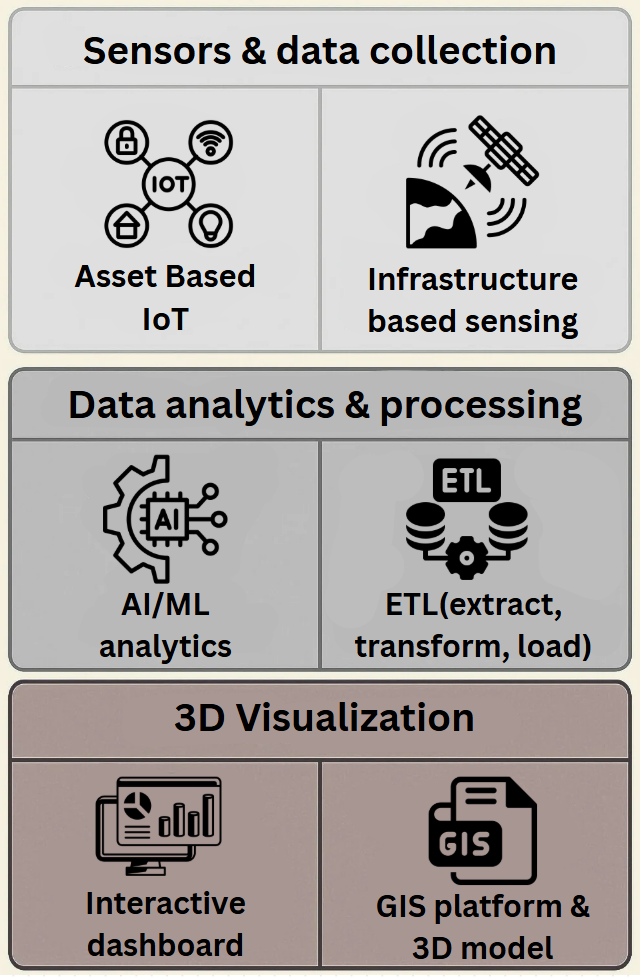}
    \caption{Overview of the DT use case for disaster management }
    \label{fig:twin}
\end{figure}

 Digital cities also facilitate the way management communicates with citizens during emergencies \cite{pang2020collaborativecitydigitaltwin,ORBi-07258921-b59d-4de9-bb00-0e1a41a578c4,Adr}. The data for the city is continuously updated, and it can be made available as live public data (e.g., through apps, websites, or social media) to inform citizens about actions. Many smart city systems ingest official sensor data, crowd-sourced inputs, and social media signals to create a two-way information flow \cite{WHITE2021103064}. This enables more localized and timely public warnings and instructions. For example, a city's digital platform might push notifications about which neighborhoods should evacuate while also ingesting tweets or smartphone reports from citizens on the ground \cite{Falco}. Research indicates that a DT Smart City can give the public and decision-makers highly localized data about a developing disaster. By keeping citizens informed (and sometimes engaging them through reporting), digital cities enhance the community's ability to respond and prevent the spread of misinformation during emergencies \cite{SHAHBAZI2024102780}.

  Figure \ref{fig:twin} illustrates the use case of DT for disaster management; this digital-city disaster management framework features a three-level architecture. At its foundation, the Sensors \& Data Collection layer brings together asset-based IoT devices (e.g., building sensors, smart meters) and infrastructure-level sensors (e.g., satellites, ground-based radar) to aggregate real-time environmental and structural information. Such live data streams are then processed in the Data Analytics and processing layer, where AI/ML algorithms identify anomalies, and ETL pipelines scrub, integrate, and normalize multisource feeds to take action. Towards its upper end, the 3D Visualization layer provides operators with interactive dashboards and GIS-enabled 3D models, enabling them to view live conditions, simulate “what-if” scenarios, and coordinate resources with pinpoint accuracy—thereby closing the loop between sensing and decision-making in emergencies. With these capabilities, digital cities can better manage disasters.

\subsection{Bidirectional Digital Twin: Concepts and Architectures}
\label{sec:digitaltwins}

A defining characteristic of a true DT is bidirectional connectivity between the physical object and its virtual model. The DT continuously ingests data from the physical asset (e.g., through IoT sensors) \cite{app142412056}. In turn, the insights or outputs from the DT can be fed back to influence the physical asset's operation \cite{app142412056}. For example, A DT of a building might receive information on temperature \cite{menges2024predictivedigitaltwincondition} and occupancy and, based on that, adjust the building's HVAC settings through control signals and create a feedback loop \cite{Hoas}. As the physical system changes, the DT changes \cite{AROWOIYA2024641}; and When the situation within the digital counterpart of a DT indicates that action is necessary, the intelligence created from the analytics of the digital counterpart can be used to initiate a response in the physical system. In a word, the DT is a living virtual model; it provides a continuous, up-to-the-moment representation of a physical asset or process that enables constant monitoring, analysis, and even control of the asset in ways the static models of the past could never hope to provide \cite{Boschert2016}.

Enabling a bidirectional DT requires an architecture that facilitates continuous data exchange and feedback \cite{robinson2024twinetconnectingrealworld,El-Agamy2024}. In general, such an architecture contains several layers. At the foundation, sensors and IoT devices on the physical asset or environment collect real-time data (measurements, status signals) and transmit them using wireless networks or the internet to feed into the DT platform \cite{CHeraku,Piya}. The incoming data is processed and integrated into the asset's virtual model. The DT platform (which may be hosted on cloud servers or edge computers) continuously updates the virtual model's state. Advanced analytics, such as ML/DL techniques \cite{yang2025leveraginglargelanguagemodels,ijgi13120445}, might be applied at this stage to analyze the data, run simulations, or predict future states. Based on the updated model and analyses, the DT generates outputs – alerts, optimization decisions, or direct control commands. In a fully bidirectional setup, specific outputs are sent back to the physical asset or to actuators in the environment to adjust operations. For example, a DT of a power grid might predict an overload and trigger a load-balancing response by routing electricity differently \cite{Othman,YASSIN2023100039}. In many real cases, the "actuation" is indirect – the DT provides decision support for human operators, who then implement actions in the physical world \cite{mylonas2024facilitatingaioperatorsynergy}.

Traditional monitoring systems often stopped at unidirectional data reporting. By contrast, a bidirectional DT continuously cycles information \cite{s23052659}. A clear illustration of bidirectional DT is in next-generation early warning systems. Conventional early warning collects sensor data and then issues warnings manually. With bidirectional DT early warning, live sensor data is inputted to a digital city model \cite{LAGAP2024104629}. The system will automatically warn or trigger protective measures if it recognizes the danger. Such systems can autonomously adapt and respond to evolving risks in disaster management \cite{Yang2024}. The applications range from micro-level (e.g., an individual building’s DT helping firefighters navigate it safely) to macro-level (a whole-region DT modeling the spread of wildfires) \cite{Seti,Bartos2022,Mankowski2020}.

\subsection{Wild Fire Disaster Management Using DT}

New advances in DT technology for wildfire management have demonstrated capabilities for near real-time simulation, predictive modeling, and sensor integration. However, several significant gaps remain in developing a fully bidirectional DT—i.e., not merely mirroring the actual environment but actually offering field intervention.

Many state‑of‑the‑art architectures—such as the NGFR hub \cite{Shaposhnyk} and the forest DT architectures described in \cite{Doll} and \cite{gii} —are based primarily on unidirectional data flows. In these systems, real‑time sensor data is used to update digital models, but the models rarely "close the loop" by influencing physical operations or modifying operating decisions based on the results of the simulations. Although approaches such as the latent data assimilation technique \cite{nhess-23-1755-2023} update analytics models in near real-time, they function primarily as basic unidirectional DTs. A genuine bidirectional DT would allow for a closed-loop architecture of monitoring and actions where the real-time data analytics initiate interventions (e.g., re‑routing UAVs or alerting first responders) and provide continuous updates to the DT.

Also, several studies \cite{LAGAP2024104629,fire7110412,10155819} have shown improvements in integrating remote sensing data with UAV and ground sensor network data. Most applications concentrate on a single data aspect—i.e., prediction of the time of burning or the global area burned—without reaching seamless unification of disparate data sets. Forest DTs must integrate disparate sources (high-resolution remote sensing and even first responder cognitive workload monitors) \cite{xiao2023estimatingdurationusingregression}.

Also, Works presented by \cite{nhess-23-1755-2023,Guzman} center on large-scale data collection with high-resolution remote sensing and IoT sensor networks. Data latency, integration mismatches, and visualization lags persist. However, Minor lags or discrepancies will desynchronize the real and virtual spaces in extremely dynamic wildfire scenarios. Although latent data assimilation methods help minimize the accumulation of errors, resilient bidirectional interfaces are still needed to offer real-time high‑fidelity synchronization and easy-to-understand visualization for rapid decision‑making. 

Ultimately, we can note that the use of Agentic AI enables us to address numerous existing problems in wildfire disaster management. We will explain the use of Agentic AI in the following sections.

\subsection{AI Agent Architectures and Approaches}

AI agents deployed in disaster management can span a spectrum of architectures, each suited to different tasks and decision paradigms. The major categories include rule-based agents, reinforcement learning agents, multi-agent systems, and large language model-based agents. Often, a comprehensive system will combine aspects of several types (for example, a multi-agent system may use RL-trained agents and an LLM-based interface). Below, we provide an overview of each category and their roles in the domain of disaster management.

\subsubsection*{Rule-Based Agents (Expert Systems)}

Rule-based agents were some of the first AI support tools utilized in emergency management \cite{IGI}. These agents are based on if-then rules or expert-designed logic and store domain knowledge and organizational procedures. In a disaster scenario, a rule-based system would automatically raise an alert when a sensor reading crosses a threshold and applicable conditions are present. Emergency managers can incorporate rule-based modules into alerting systems (for discrete triggers) and evacuation procedures (e.g., when the fire crosses Zone X, evacuate Zone Y). However, rule-based agents are not highly adaptable. Disasters may introduce new situations or nuanced trade-offs that formal static rules cannot cover. Nevertheless, rule-based elements are still a fundamental component of hybrid AI designs, typically acting as a safety mechanism or for regulatory reasons (implementing policies and checklists that need to be followed). In summary, rule-based agents provide a foundation of domain knowledge and reliability; however, they require ongoing updates and expansion by domain experts.

\subsubsection*{Reinforcement Learning Agents}

Reinforcement learning can be used as an efficient means of developing disaster management agents to learn policies optimally by simulated trial and error. In reinforcement learning, the agent repeatedly makes decisions within an environment, receives feedback in the form of rewards or penalties, and learns policies over time to maximize long-term rewards (for instance, the lives saved and the mitigated damage).

The paradigm is particularly well-suited to dynamic uncertainty problems, which are typical of disaster situations \cite{zhou2024realtime}. By experimenting with varied sequences of actions within the simulation, RL agents can identify innovative solutions that human planners may not consider, particularly when the problem space (e.g., every possible evacuee route and time for thousands) is large and complex. For example, in one study, a deep Q-network agent was employed to manage evacuations in buildings, and policies learned had more than 90\% successful rescue rates under simulations, which significantly outperformed heuristically based solutions \cite{zhou2024realtime}. The benefits of RL agents are their ability to learn and sequentially decide in complex decision-making processes, as well as not requiring explicit environmental modeling.

\subsubsection*{Multi-Agent Systems and Collective AI}

Many disaster management problems inherently involve multiple agents by nature, as numerous responders,  vehicles, and devices all act and interact. Multi-agent systems (MAS) provide an architectural framework for designing AI, where multiple intelligent agents pursue goals (sometimes shared, sometimes individual) within a familiar environment. We have already discussed several MAS applications in the coordination section; here, we highlight the architectural and algorithmic perspective of multi-agent AI in disasters.

In a MAS, each agent may be a relatively simple entity (such as software or a robot), but significant emergent behavior arises from their interactions. Agents can be homogeneous (all of the same types, as in a swarm of identical drones) or heterogeneous (e.g., a mix of sensors, robots, and decision software, each with different roles) \cite{smythos2025utilizing}. Key to MAS is the communication and negotiation protocols that allow agents to coordinate. Distributed planning algorithms can be employed to enable agents to agree on task assignments or allocate resources without a central commander.

One cutting-edge area is multi-agent reinforcement learning (MARL), which combines the learning paradigm with agent collectives. MARL enables agents to learn not only how to optimize their actions but also how to cooperate or compete with other agents to achieve system-wide objectives \cite{smythos2025utilizing}. In disaster simulations, MARL can be used for teams of heterogeneous agents, e.g., a team with drones (for search) and ground vehicles (for rescue) learning a joint policy where drones learn to efficiently cover search areas and signal the ground team and ground agents know where to reposition in anticipation of drone findings.

The advantages of MAS are evident in their modularity, scalability, and fault tolerance. As tasks grow, more agents can be added; if an agent fails, the system degrades gracefully rather than catastrophically. By specialization, each agent can focus on a sub-problem (surveillance, transport, logistics, communication). Moreover, MAS can naturally mirror the organizational structures of emergency response (which are often distributed and multi-tiered) \cite{chen2024promise}. 

\subsubsection*{Large Language Model-Based Agents}

Over the past few years, large language models have significantly enhanced the capabilities of AI to process and generate human languages. LLM-based agents differ from traditional expert systems or control agents in that they excel at communication, information synthesis, and knowledge retrieval. This makes them ideally suited for roles that involve interfacing between humans and vast information sources during times of crisis. For example, Researchers introduced the integration of an LLM  into the 911 emergency dispatch process \cite{Khan}. In this setup, whenever a 911 call is received, a speech recognition module first transcribes the caller's spoken words in real-time.

The LLM agent evaluates the call text in parallel with the human dispatcher, performing tasks such as extracting critical facts (location, nature of the emergency, and number of people involved), as well as determining the caller's language or any communication barriers. The LLM can serve as an assistive tool to the dispatcher, providing follow-up questions to ask the caller (useful when the caller is panicking or giving incomplete information). In multilingual applications, the LLM can translate a caller's speech when the caller is not an English speaker and can recommend short phrases for the dispatcher to communicate, essentially functioning as a live translation and coaching tool.

For the general public, large language-model-based chatbots can serve accessible crisis information and instructions. In the event a wildfire is headed towards a community, a publicly accessible chatbot (web-based or SMS-based) can utilize an LLM to respond to residents' questions, such as "What do I need to evacuate?" or "Where is the closest pet-accepting shelter?" LLM-based agents can handle broad question sets and provide context-based suggestions, following up on follow-up questions as well in a conversational tone. In overpowered call centers during disasters, such AI assistants can serve as information triaging, providing people with timely responses.

\subsection{Situation Room in Crisis Management}

A situation room (also called a command center, operations center, war room, or Emergency Operations Center (EOC)) is a dedicated facility or digital environment where decision-makers collect, integrate, and analyze real-time information to coordinate actions during crises. In practice, it is the nexus of command and control for emergency management or security operations. As FEMA notes, an EOC is a central command and control system responsible for emergency management at a strategic level from which leaders coordinate information and resources to support incident management \cite{FEMA2020}. Similarly, Herrera et al. define a situation room as a physical or virtual space where experts systematically analyze information to characterize an evolving situation \cite{Herrera2021}.

Situation rooms integrate specialized hardware, software, and human resources in a configurable environment. Hardware typically includes ergonomic consoles and workstations for operators, high-resolution video-wall displays, and reliable communications equipment (secure networks, satellite uplinks, radio links). Large multi-screen video walls are ubiquitous – no command center is complete without a video wall system – because they simultaneously present multiple data streams (e.g., live surveillance feeds, maps, sensor dashboards, alerts, and communications) to enhance situational awareness \cite{ConstantTech2025}. Operators may also use augmented-reality tablets or mobile devices to receive and forward information from field units. On the software side, situation rooms run integrated platforms that aggregate data from diverse sources: geographic information systems (GIS), environmental sensors, utility grids, social media feeds, public safety dispatch systems (911 calls), and weather feeds. These platforms typically implement a service-oriented architecture or publish/subscribe messaging to fuse real-time data into a unified dashboard \cite{Fannnnggg}.

The primary functions of a situation room (physical or virtual) include collecting, analyzing, and sharing information; tracking and allocating resources; coordinating response plans; and sometimes, issuing policy or tactical directives. During a crisis, the situation room acts as the coordination hub: analysts from different agencies report incoming data, update status boards, and quickly surface critical issues. By bringing together public safety, utility, health, and other stakeholders, the situation room ensures that responders in the field and leaders in the center share a synchronized view of unfolding events.

Situation rooms are widely deployed in government and enterprise. At the national level, executive headquarters (e.g., the U.S. White House Situation Room) and military command centers fit this model. Cities build Integrated Operations Centers to coordinate public safety; for instance, smart-city control rooms aggregate traffic cams, 911 feeds, and utility data \cite{Fannnnggg}. Private sector security also uses SOCs (Security Operations Centers) with similar features \cite{ConstantTech2025}. Other domains use similar concepts: banking institutions may have war-room dashboards for cybersecurity breaches, and companies leverage executive war rooms to manage crises (e.g., supply-chain disruptions). The pattern is the same in all cases – integrate all pertinent information, present it to experts, and coordinate the appropriate response.

\subsection{Gap in the Literature}

Numerous studies  \cite{Shaposhnyk,gii} have begun addressing decision support, such as the monitoring of the cognitive workload of emergency responders, but there remains a gap between on‑site operational control and high‑fidelity simulation models. The gap is the unavailability of an integrated, bidirectional decision‑support system translating the outputs from the DT (e.g., risk maps, forecasted fire spread) into executable instructions for both automated and human operators by utilizing AI agents and intelligent systems. It is necessary to fill the gap by incorporating the findings in studies carried out by  \cite{Zhouuuu,Guzman} to have adaptive control through real‑time feedback.

Moreover, there are specific issues despite significant interdisciplinary advances—from machine learning and reduced order modeling to IoT and remote sensing—the literature \cite{fire7110412,LAGAP2024104629} shows that standardized protocols for interoperability between heterogeneous systems are still at the nascent stage. Framework development and communication protocols that enable barrier‑free two‑way data exchange are crucial to have feedback from the DT to continuously trigger physical action. This standardization is one significant research gap our project addresses directly.

Despite the proliferation of machine–learning–based anomaly detectors and predictive fire–spread models, the role of AI in current digital-twin frameworks remains largely advisory rather than operational. Most AI modules function as isolated “black‐box” components—flagging risks or forecasting scenarios—but they are neither orchestrated by a central intelligence nor granted authority to dispatch interventions. Reinforcement-learning and multi-agent coordination techniques have shown promise in simulated environments; however, they have not been integrated into a unified, agent-mediated command layer that can learn continuously from live data, match unfolding events to pre-computed strategies, and issue executable control commands. Even large‐language‐model interfaces, which could synthesize heterogeneous streams and facilitate human–AI collaboration, are seldom tapped for real‐time decision orchestration. Without embedding AI agents directly into a closed‐loop Intelligent Situation Room—capable of adaptive planning, transparent reasoning, and seamless human oversight—digital twins will continue to fall short of delivering truly proactive, automated wildfire response.

 Based on previous parts and the above explanations, integrating bidirectional feedback into DT systems for wildfire management is not yet mature. To close these critical gaps, we propose embedding an agent‐mediated Intelligent Virtual Situation Room (IVSR) on top of the bidirectional DT core. At its heart, a standardized, publish/subscribe data bus unifies all sensor, GIS, and simulation streams into a common schema, while AI agents—ranging from reinforcement‐learning dispatchers to LLM‐based orchestration modules—continuously monitor live DT outputs (risk maps, spread forecasts, resource status) and match them to precomputed intervention strategies in the Disaster Simulation Library. Once an AI agent selects or refines a plan, it automatically generates executable commands—whether to reroute UAVs, reallocate crews, or trigger building systems—and dispatches them through the same standardized protocols back into the physical layer, all under a lightweight expert‐approval workflow. By combining end‐to‐end interoperability standards with a closed‐loop, learning‐enabled command layer, our IVSR transforms advisory analytics into real‐time, adaptive control, thereby delivering the first truly operational, bidirectional DT solution for wildfire management proven by simulations from the company case.

\section{Methods \& Materials}

\begin{figure}[ht]
    \centering
    \includegraphics[width=\columnwidth]{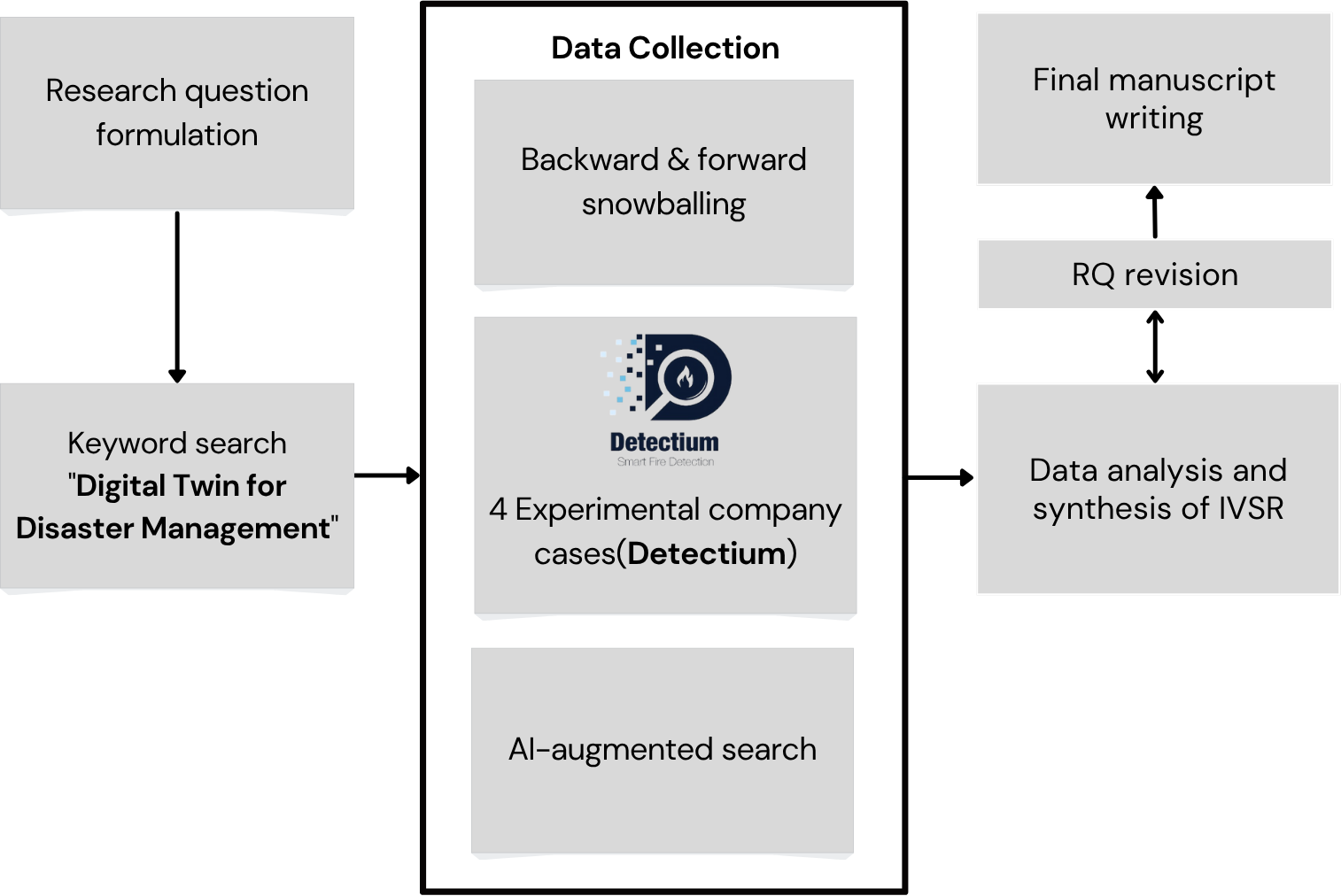}
    \caption{Research Method Structure}
    \label{fig:Method}
\end{figure}

We used primary and secondary data including a number of cases from a DT-based fire detection company as well as existing examples of the technology deployment from a literature review. One of co-authors is from the case company who provided the case details and technical data related to the implementations. The literature review focused on publications mostly from 2023 through 2025, retrieved via Google Scholar to capture the most recent advances in DT technology for disaster management. Recent reviews on wildfire DTs revealed that the search strategy was designed to ensure comprehensive coverage of relevant studies. The following steps were executed in the literature search: We used "Digital Twin for Disaster Management" to retrieve candidate papers from Google Scholar. After identifying key papers, we performed backward and forward snowballing by examining these articles' reference lists and citation links \cite{Wohlin}. This iterative technique helped us find additional relevant studies that may not have appeared in the initial search. To capture any semantically related works not linked by citation, we input the titles and abstracts of selected papers into a GPT-based large language model. This AI-assisted search yielded further candidate studies by identifying related keywords and topics in an expanded corpus.
This AI-assisted search yielded further candidate studies by identifying related keywords and topics in an expanded corpus.
This three-pronged approach—keyword searching, snowballing, and AI-augmented search—guaranteed comprehensive literature coverage. Retrieved papers were screened using predefined criteria. Inclusion criteria stipulated that a paper (1) discussed DTs in a disaster or wildfire scenario, or (2) employed real-time data fusion and dynamic simulation with the use of AIs, and (3) supported bidirectional feedback between physical systems and the DT. We favored studies with practical implementations or case studies. These criteria capture the core properties of DTs (e.g., bi-directional information interaction) and the focus of real-time simulation in wildfire scenarios. Figure \ref{fig:Method} illustrates the structure of our research method.

The experimental part consisted of case-study simulations of fires in industrial settings. Detectium, a technology company that offers digital-twin solutions for fire safety, provided realistic scenarios. Each scenario represented a DT system with the facility's layout. The DT platforms were set up to mimic the fire's spread and the systems' responses over time.

\section*{Analysis of the Company Case}
In this section, we present the results of our case studies and synthesize their usefulness for a broader deployment of an intelligent virtual situation room concept.
The case company, a Finland-based deep-tech startup, utilizes a digital twin combined with computer vision for the early detection and prevention of fires in industrial and agricultural facilities. They use 3D visualization to show the facility's status regarding the fire in real-time to relevant stakeholders. This method is interesting because it combines data collected from two different sensors, as well as third-party weather data, into a cohesive and real-time image of a facility or warehouse regarding fire anomalies. 

\subsection{Sensors 3D Coverage Optimization }

After setting up the virtual 3D scene for the DT, the camera coverage calculation algorithm in Unity is used to optimize sensor placement and angle, maximizing the effective detection coverage area. This optimization process enhances the fire detection system deployment costs and balances it with highest coverage, while taking into account the 3D features of the space. Moreover, this accurate coverage estimation using the 3D model of the DT enables consideration of the equipment and machinery within the environment. It provides a way to maximize fire detection coverage using a minimal number of sensors, while reducing sensors' coverage overlaps during monitoring and lowering the cost of system implementation and installation activities.

\begin{figure}[ht]
    \centering
    \includegraphics[width=1\columnwidth]{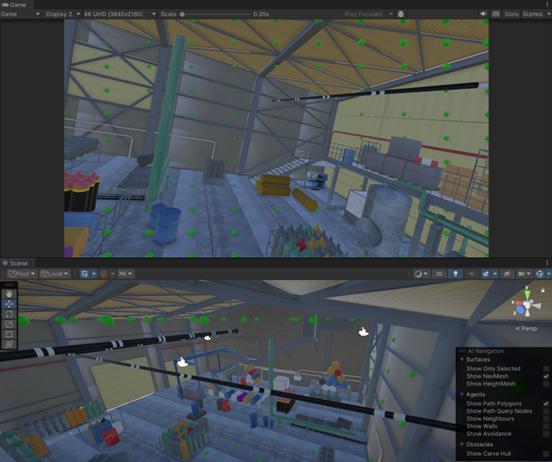}
    \caption{Visualization of intersections in the scene from 100 rays emitted by the virtual camera.}
    \label{fig:camcover1}
\end{figure}
\begin{figure}[ht]
    \centering
    \includegraphics[width=\columnwidth]{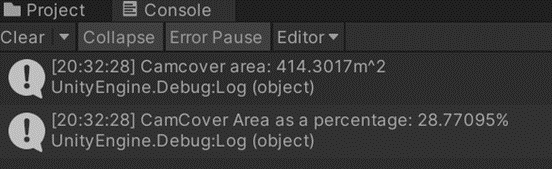}
    \caption{Calculated camera coverage area as a percentage.}
    \label{fig:camcover2}
\end{figure}

This algorithm calculates the coverage area of the virtual camera (corresponding to real-world sensors) and its proportion to the total detectable area in the scene, allowing for adjustments to the sensor's position to achieve the best coverage. The specific calculation steps are as follows:
\begin{enumerate}
    \item Assign collision attributes only to the floor, walls, and ceiling in the scene. Disable collisions for other objects if present. Calculate the total area of the floor, walls, and ceiling, denoted as $S_0$.
    \item Load the camera coverage calculation algorithm, referencing the targeted virtual camera.
    \item After triggering the calculation, the algorithm will create a $10\times10$ grid of points using the virtual camera's view, shooting 100 rays across the scene and visualizing coverage areas in 3D space with spheres at the collision points.
    \item Calculate the total area of the shapes formed by these intersections to determine the coverage area $S_1$ of the virtual camera and the percentage of total area covered $P_1 = S_1/S_0$. Log and display $S_0$ and $P_1$ values to the user.
\end{enumerate}
Figure~\ref{fig:camcover1} illustrates the visualization of intersections from 100 rays, and Figure~\ref{fig:camcover2} shows the calculated camera coverage area along with its percentage.

\subsection{Fire Localization in 3D Model }

Early and localized detection is crucial to address anomalies that can result in a catastrophic fire. Utilizing DTs to visualize and localize fire incidents in real-time may enable directed and even automated responses with less damage to the surrounding areas. 

Incident localization relies on an accurate and updated 3D environment model of the DT. To achieve fire localization the case company developed a code to map the detected 2D flame coordinates from the sensor image and transfer it into the 3D virtual scene using virtual cameras. These virtual cameras are positioned in the 3D model according to the actual sensor positions and orientations in the real world. 

\begin{figure*}[ht]
    \centering
    \includegraphics[width=1\textwidth]{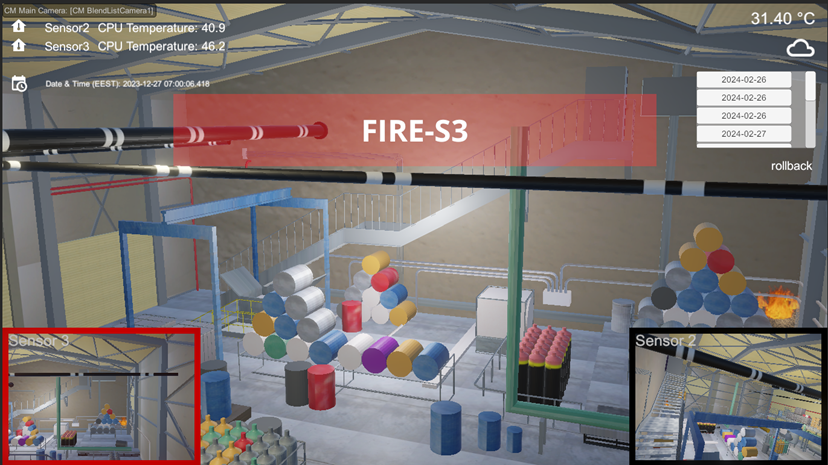}
    \caption{Example of flame location points in the virtual camera and the generated flame position in the scene.}
    \label{fig:incident_localization}
\end{figure*}

In Unity, receiving sensor fire detection data and localizing it in a virtual scene involves four steps related to mapping sensor data onto a virtual camera and raycasting in a 1:1 virtual scene with the real one. The process is as follows:
\begin{enumerate}

    \item Sensors send JSON strings containing fire detection, temperature, and location data to Unity.
    \item Unity maps the location data onto the virtual camera's 2D view upon confirming fire presence.
    \item A virtual camera, matching the real-world sensor's location and settings, receives this data and casts a ray from the mapped point in the scene, creating a fire prefab at collision points.
    \item Fire alarms are displayed on the virtual camera. If multiple sensors detect fire, the same raycasting process visualizes flames in the scene (viewable from the main perspective), with details stored in a rollback file.
\end{enumerate}
Figure~\ref{fig:incident_localization} demonstrates an example of flame location points in the virtual camera and the corresponding generated flame position in the scene.

\subsection{Incident Rollback \& Anonymization of Incident Footage}

Employing DTs to roll back in time, enabled by the incident log, compresses footage of fire incidents for efficient storage and future analysis. Each detected fire incident's information, such as size and coordinates, is saved in a file, sent to Unity, and displayed in a scroll view in the user interface. Moreover, using 3D anonymized avatars protects the privacy of individuals captured in the footage while maintaining the integrity of the data.

Each time the sensor detects a fire, the current detection parameters are saved as a \texttt{.txt} file log. An example entry is as follows:
\begin{table}[htbp]
\centering
\caption{Example JSON Log Entry}
\label{tab:jsonlog}
\begin{minipage}{\columnwidth}
\begin{lstlisting}
{"FireThreatLevel": "probable fire", "StartDateTime": "2024-03-17 07:05:36.137095", 
 "CPUTemperature": 50.1, "SensorId": "", "Column": "107", "Row": "67", 
 "Temperature": "107", "Number": "400"}
\end{lstlisting}
\end{minipage}
\end{table}
Here, \texttt{FireThreatLevel} indicates the severity of the fire threat, \texttt{StartDateTime} the time the fire was detected, \texttt{CPUTemperature} the sensor's CPU temperature, while \texttt{Column} and \texttt{Row} denote the coordinates of the highest temperature pixel in the sensor’s visual field. The \texttt{Temperature} field shows that pixel’s temperature and \texttt{Number} the total pixel count detected for the target flame.

\begin{figure*}[ht]
    \centering
    \includegraphics[width=1\textwidth]{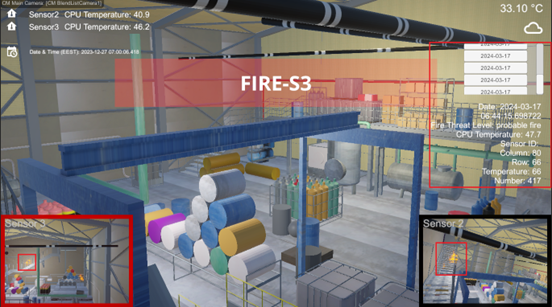}
    \caption{An example of fire generated in the scene when clicking on a record in the rollback list.}
    \label{fig:rollback}
\end{figure*}

Each fire detection record is saved in a \texttt{.txt} file, which is sent to Unity and displayed in a scrollable user interface. Clicking an entry in the rollback log triggers Unity to resend this information to the virtual camera, reenacting the above steps to generate a fire event that lasts for 30 seconds. This enables users to review the location and details of all historical fire alerts in the 3D environment.
Figure~\ref{fig:rollback} shows an example of a fire generated in the scene after clicking a record from the rollback list. Additionally, as shown in Table~\ref{tab:jsonlog}, which is generated after clicking a record from the rollback list, each log entry contains detailed sensor readings, including fire threat level, timestamp, temperature, and pixel data, which are crucial for reconstructing and analyzing fire events.

\subsection{Fire Spread Simulation \& Customization of Machine Learning Models}

\begin{figure*}[ht]
    \centering
    \includegraphics[width=1\textwidth]{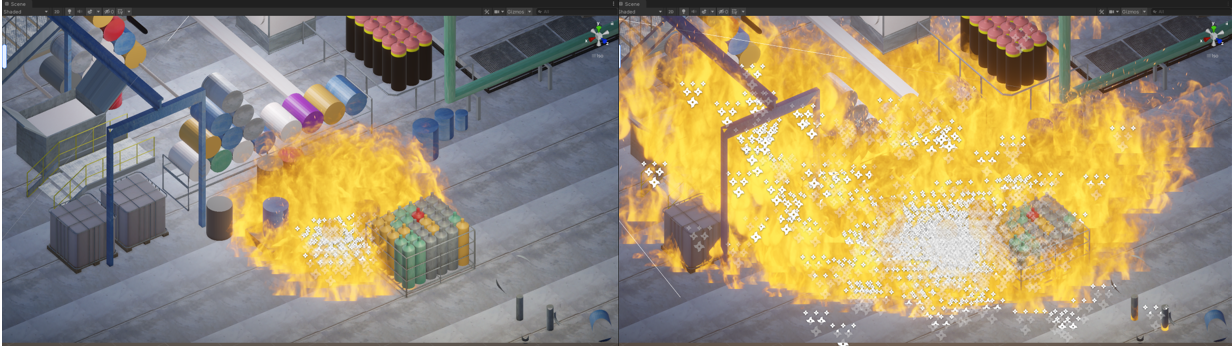}
    \caption{Example of the initial stage of flame generation and its expansion after a period of time.}
    \label{fig:flame_expansion}
\end{figure*}

Another use case of the dimensionally accurate 3D model of a DT is in fire propagation simulation, where the material types of different elements within the model are defined. In this use case, the direction and rate of propagation of an incident can be visualized in any scenario, and based on the observations, escape routes, and rescue procedures can be established. Leveraging DTs to simulate fire scenarios before incidents occur enables the testing and refinement of emergency response procedures and strategies. By analyzing various scenarios, firefighting teams can better prepare for different types of fire events and improve their response capabilities. Fire propagation simulation during an ongoing incident can enable data-driven rescue operations and decision-making. 
 The flame expansion simulation is based on the expansion of a spherical collision body centered on the flame and calculates its intersection points with other object collision bodies in the scene. Due to the absence of complex particle calculations, compared to other Unity flame combustion and spread tools such as Ignis \cite{Arctibyte2024Ignis}, the flame expansion simulation algorithm in this article has lower performance overhead and can run on the WebGL web platform. In this method, the specific steps of flame expansion simulation are as follows: 
\begin{enumerate}
    \item Set up mesh colliders for objects with different burning properties in the scene, set all colliders as triggers, and label each object with different material tags as criteria for determining the burning rate during flame spread.
    \item Generate an empty object coordinates on the positioning point of the flame in the virtual scene, attach a spherical collider to this object, and set the expansion speed of the spherical collider based on mappings of relevant sensor data fields.
    \item As the spherical collider gradually expands, calculate the intersecting area between the spherical collider and colliders of other objects in the scene in each frame. Based on different material tags of objects, generate new flame prefab clones at these intersecting areas at different rates.
    \item Real-time adjusts the position coordinates and expansion speed of the flame spread collider based on new fields transmitted by sensors. Display the time of flame spread and three-dimensional simulation information on the GUI interface for the user and provide corresponding action recommendations.
\end{enumerate}
Figure~\ref{fig:flame_expansion} demonstrates the initial stage of flame generation and its subsequent expansion over time.

Additionally, the DT can generate synthetic data that is then used to retrain generic machine vision models to improve their performance on the DT site. In other words, by customizing the machine learning models for individual DTs, we can improve each site's fire and anomaly detection and predict fire events based on specific environmental factors and contexts. This tailored approach enhances the accuracy and reliability of fire detection systems for each DT \cite{khajavi2024synthetic}
.

\subsection{ Physical Intervention \& Issuing Commands }

The case company is currently developing a system that leverages a digital twin to enhance incident response capabilities. The digital twin serves as a real-time 3D representation of the facility, providing situational awareness by visualizing fire incidents as they occur. This model is integrated with live data from drones operating within the facility, including their current location and operational status. In a bidirectional setup, the digital twin not only receives real-time input from the drone but also sends control commands. Upon detecting a fire, the system calculates an optimal 3D route for safe and rapid drone deployment to the incident site. This enables the drone to respond quickly and efficiently for early fire extinguishment. The integration of real-time data and autonomous control through the digital twin represents a novel approach the company is actively exploring.

\begin{figure}
    \centering
    \includegraphics[width=1\columnwidth]{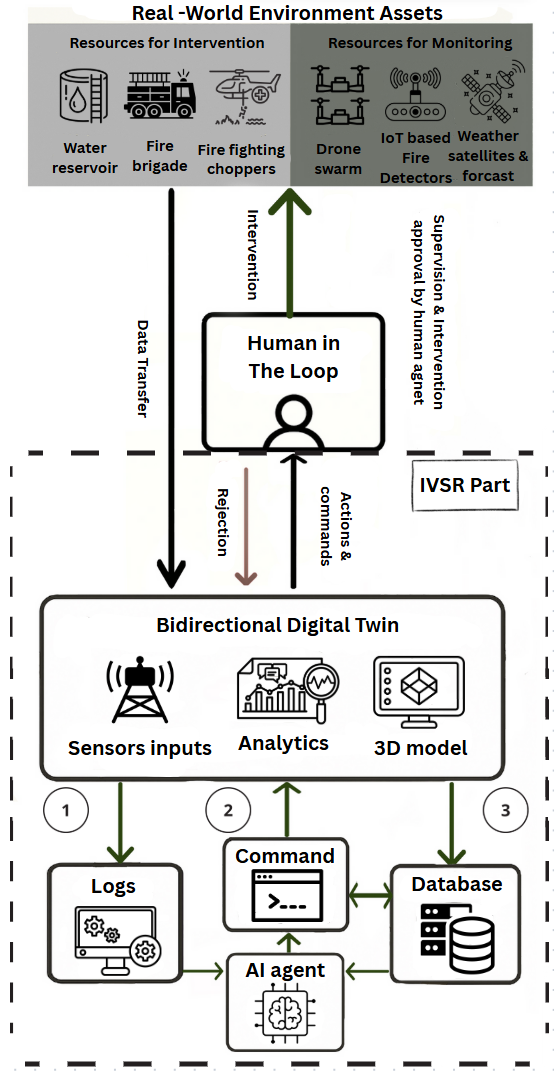}
    \caption{IVSR architecture. (1) Real-time status parameters from the Digital Twin at the moment of the incident are streamed into the system, conveying up‐to‐the‐second environmental, asset, and resource information. (2) The AI agent uses these live parameters to retrieve the most relevant response plan from the pre‐computed Simulation Library and fine‐tunes it based on current conditions. (3) The Simulation Library contains a collection of DT scenarios and associated intervention strategies generated before any real event.
}
    \label{fig:IVSR}
\end{figure}

\subsection{The Conceptual Architecture of Intelligent Virtual Situation Room}

Through the review of the literature and analysis of primary data related to the technical feasibility of a bidirectional digital twin, we synthesize an architecture for an IVSR which is a decision and intervention-support system designed for real-time disaster management.
 
 The IVSR, presented in Figure \ref{fig:IVSR} is powered by a bidirectional DT, which enables continuous data collection, monitoring, and command-and-control operations. At the heart of the system lies a Disaster Simulation Library (DSL)—a comprehensive database containing pre-simulated wildfire scenarios and associated intervention strategies. These simulations account for both controllable and uncontrollable variables, including wind speed, humidity, fire suppression resources (e.g., firefighters, hydrants, aerial units), water levels in reservoirs, and other relevant environmental and operational parameters. Each scenario in the DSL documents the detailed deployment and response tactics used to contain the wildfire. Moreover, interventions are ranked based on their effectiveness, cost-efficiency, and response speed, covering a wide spectrum—from routine, low-impact events to rare, high-consequence incidents.

During an active event, the DT ingests real-time data from the field to generate a live status log of the environment and available response assets. An integrated AI agent analyzes this log to identify the most similar scenario from the DSL, retrieving corresponding intervention plans. The status log is continuously updated to refine the similarity assessment as the situation evolves, ensuring adaptive response recommendations are provided. Once similarity is validated, the intervention commands are scheduled and dispatched to field units, enabling rapid and coordinated action—minimizing confusion and delays that often characterize the critical early hours of a disaster. To ensure safety and reliability, all commands are routed through a panel of supervising experts via the DT for approval before implementation. Approved commands are executed, and their outcomes are fed back into the system for real-time monitoring and further decision support. If a command is rejected or modified by the panel, the IVSR logs these changes continuously by its DT to improve its recommendations. 

The IVSR delivers two primary benefits: (1) seamless integration of diverse asset and environmental data for improved visibility and situational awareness, and (2) faster, more confident, and semi-automated orchestration of human and autonomous response systems—ensuring timely, effective, and informed disaster response.

To further define the architecture of the Intelligent Virtual Situation Room (IVSR) and describe how parameters like the quantity of available first responders, fire severity, resources available,  and more can be incorporated and integrated, the system can be dissected into its parts and described as how and what data it ingests, processes, and leverages to inform decision-making in near-real time. The fundamental functionality of the IVSR is to create a real-time virtual model of the environment in which the wildfire is happening. The environment is constantly updated with real-time information from multiple sources. Using this information, the IVSR provides an interactive 3D virtual environment for decision-makers to monitor and manage the response to the wildfire.

\begin{table*}[htbp]
\centering
\caption{JSON Log Entries of Fire Incident Sensor Data in Several Different Days}
\label{tab:jsonlog}
\begin{lstlisting}
{"FireThreatLevel": "fire hazard", "StartDateTime": "2024-02-26 10:25:54.334260", "CPUTemperature": 49.1, "SensorId": "", "Column": "76", "Row": "39", "Temperature": "71", "Number": "1007"}
{"FireThreatLevel": "probable fire", "StartDateTime": "2024-02-27 21:40:09.074818", "CPUTemperature": 56.0, "SensorId": "", "Column": "83", "Row": "50", "Temperature": "72", "Number": "1073"}
{"FireThreatLevel": "probable fire", "StartDateTime": "2024-03-04 04:46:13.202399", "CPUTemperature": 50.1, "SensorId": "", "Column": "79", "Row": "79", "Temperature": "69", "Number": "315"}
{"FireThreatLevel": "probable fire", "StartDateTime": "2024-03-10 04:17:32.398583", "CPUTemperature": 49.1, "SensorId": "", "Column": "83", "Row": "73", "Temperature": "78", "Number": "412"}
{"FireThreatLevel": "probable fire", "StartDateTime": "2024-03-17 05:52:32.027040", "CPUTemperature": 47.2, "SensorId": "", "Column": "97", "Row": "58", "Temperature": "148", "Number": "626"}
{"FireThreatLevel": "probable fire", "StartDateTime": "2024-03-17 06:44:15.698722", "CPUTemperature": 47.7, "SensorId": "", "Column": "80", "Row": "66", "Temperature": "66", "Number": "417"}}
\end{lstlisting}
\end{table*}

\subsubsection*{Key Parameters in the IVSR}

The IVSR can monitor the deployment and availability of first responders, including police officers, paramedics, and firefighters. They are depicted in the virtual environment based on their status and their exact locations. For instance, if a wildfire has swept across several kilometers, the IVSR will display the locations of responders closest to the fire, the resources available (e.g., fire engines, water tanks), and their current tasks. Commanders can use this information to decide where to send reinforcements. 

The IVSR monitors several parameters that define fire severity through sensors, including fire intensity, rate of spread, and temperature. The data is continuously fed into the virtual model. Suppose the fire is intensifying in a particular area. In that case, the IVSR may demarcate the region with a color-coded warning (e.g., red for high-risk), which enables emergency responders to prioritize their responses.

The system also takes into account the materials being burned in the fire. An example is forest fires, which tend to differ from fires in urban industrial situations, where various chemicals may be involved. If the fire is in a forest (with a rate of movement different from that in a metropolitan area), the IVSR could modify the simulation to factor in the fuel load (trees, grasses, etc.) and estimate the rate at which the fire is moving. Furthermore, the IVSR includes fire equipment, water outlets, aerial resources (e.g., drones or helicopters), and emergency shelter sites. The IVSR is capable of providing a comprehensive overview of all fire equipment and resources on hand and assisting decision-makers in determining when to deploy aerial suppression or ground units to a particular high-risk target area. 

Environmental conditions, such as temperature, humidity, and wind speed, are constantly tracked and utilized to adjust the simulation. The information is used to predict the future behavior of the fire. When the wind is expected to rise, the IVSR simulates how the fire could propagate more quickly, allowing decision-makers to prepare for future hotspots in advance.

\subsubsection*{How Data Is Processed and Updated in IVSR}

The IVSR gathers data from sources, processes it, and keeps the virtual space updated in real-time. Below is a step-by-step explanation of incorporating the base and new data. The data, such as readings from fire sensors, wind speed, and humidity, is collected from environmental sensors or IoT sensors. For example, the sensor network surrounding the fire will send messages such as "Temperature: 72°C, Humidity: 12\%, Wind speed: 18 km/h" to the IVSR system. All this data is incorporated into the virtual environment to dynamically update the fire simulation. 

New information, either new sensors or the latest weather forecasts, is incorporated into the system. The information is processed using machine-learning models and is adjusted during the simulation. The machine-learning-based models of the IVSR process analyze the data and dynamically change the simulation. If extra sensors are included to monitor air quality and the concentration of smoke particles, the IVSR can use this additional data to revise its predictions. If the weather forecast indicates a change in wind direction, the IVSR recalculates the potential fire propagation. The IVSR can create synthetic data from pre-simulated fire situations. Such data will be used for training decision support systems or AI systems. 

A fire situation can be simulated, for instance, where the fire has engulfed an industrial complex, and based on this simulation, provide real-time predictions of the fire and make suitable interventions. 

The table \ref{tab:ivsr_params} shows an example of how the IVSR integrates these parameters and updates in real-time:

\begin{table*}[htbp]
\centering
\resizebox{\textwidth}{!}{
\begin{tabular}{|l|l|l|}
\hline
\textbf{Parameter}                  & \textbf{Value/Status}               & \textbf{Description/Use Case}                                                                                  \\ \hline
\textbf{Fire Severity}               & High                                & The fire is spreading rapidly with a high temperature. A red alert is triggered for the area.                    \\ \hline
\textbf{Number of First Responders}  & 20 firefighters, 5 fire trucks      & 20 firefighters are available in the field, along with 5 fire trucks. Their locations are tracked on the map.   \\ \hline
\textbf{Material Type Burning}       & Forest, Dry Leaves                  & The fire is in a forest area with dry vegetation, which accelerates the spread.                                 \\ \hline
\textbf{Wind Speed}                  & 20 km/h, Direction: North           & The wind is blowing northward at 20 km/h, influencing the fire's spread towards nearby communities.             \\ \hline
\textbf{Available Resources}         & 2 Helicopters, 3 Fire Trucks, 1 Ambulance & Helicopters are available for aerial suppression, and ambulances are stationed for potential casualties.    \\ \hline
\textbf{Fire Spread Rate}            & 0.5 km/h                            & The fire is currently spreading at a rate of 0.5 km/h based on real-time data.                                  \\ \hline
\textbf{Temperature (Sensor 1)}      & 72°C                                & Sensor in the affected area reports 72°C, indicating a very high-intensity fire in that region.               \\ \hline
\textbf{Fire Object Detection}       & Detected Trees, Power Lines         & The system has identified trees and power lines in the path of the fire, critical for prioritizing response.    \\ \hline
\end{tabular}
}
\caption{An example of some IVSR parameters and their real-time implications}
\label{tab:ivsr_params}
\end{table*}

During a wildfire, as the intensity in different locations increases, the IVSR updates the simulation to project a greater risk in the locations based on the latest information. When a new fire resource is dispatched (e.g., extra firefighters or aerial vehicles), the system updates the availability of the resources and the simulation figures to reflect where the resources are needed most.

The IVSR can project possible evacuation pathways based on actual-time traffic data and provide locations where available shelters are in the case that the wildfire is forecasted to head toward a congested area. Based on the updated fire spread and available resources, the IVSR may recommend deploying fire trucks and ambulances to specific areas while sending drones to monitor the fire's progression.

Spatio-temporal features (like fire growth rate and time sequence) are also considered to match dynamic behavior. Once retrieved, the IVSR’s decision module formulates an action plan by either extrapolating from the successful responses in those past simulations or using an LLM to reason with the current and retrieved scenario descriptions. For example, suppose the IVSR detects a high-intensity and fast-spreading wildfire in a forest. In that case, it will retrieve similar wildfire incidents and likely recommend deploying aerial suppression (such as drones or water bombers) and evacuations in adjacent zones. Conversely, a localized kitchen fire indoors might retrieve analogous residential fire cases and suggest activating building sprinklers and guiding occupants to safety. These recommendations are delivered to human operators or automated AI system responders by the IVSR. When appropriate, they can trigger commands to connected devices (e.g., autonomous drones or alarm systems). The overall architecture thus continuously learns from historical fire data and employs state-of-the-art AI techniques to provide context-aware and scenario-specific response guidance in indoor and outdoor fire emergencies \cite{dhs2024aimeans}.

By constantly consuming and processing data, the IVSR enables decision-makers to possess a comprehensive, up-to-date picture of the situation in the disaster, allowing for quicker and more informed decisions and reducing the threat to human lives and infrastructure. The data flow, ever-present, not only allows for continual updating but also creates a model that enables lessons to be learned from the past and responses to be refined in the future.

\subsection{Utilizing AI Agent in IVSR}

In the proposed IVSR architecture, the AI agent plays a focused and efficient role. It continuously compares real-time parameters from the digital twin of an ongoing incident with the Disaster Simulation Library (DSL) of pre-simulated incident scenarios. The simulations in the DSL incorporate operational response variables, such as personnel and equipment availability, making them directly applicable to real-world constraints. When a high degree of similarity is detected between the current situation and one of the simulations—based on environmental conditions, damage profiles, resource availability, and response readiness—the AI agent retrieves the most relevant simulation and its associated detailed response operational strategies. The response interventions are then dynamically and rapidly presented to human operators to support timely and informed decision-making, after which the selected interventions are transmitted through the bidirectional digital twin to human-operated or autonomous units on the ground for execution.

This similarity-matching process is continuously updated as the incident evolves, ensuring that operators are always guided by the most contextually relevant past real or simulated scenarios. Unlike more complex, general-purpose AI systems that attempt to evaluate and recommend responses autonomously, our IVSR design leverages a lightweight AI agent focused solely on similarity assessment. This targeted approach enables faster, more transparent, and operationally reliable support during emergencies.

\subsection{Human In The Loop}

While the IVSR provides sufficient data and control links, the decision-making is up to the humans, as shown in Figure \ref{fig:IVSR}. Human decision-makers and AI systems each have unique strengths for disaster management. AI excels at analyzing massive information flows and detecting patterns that surpass human mental capacity \cite{Visave2025}. Still, it can also break down in unexpected ways and operate as a black box \cite{Kuglitsch2024}. Unlike humans, AI lacks intuitive reasoning, contextual judgment, and ethical discrimination, particularly in new and uncertain situations \cite{Sharma2024}. 
On the other hand, Human choice is not perfect: humans can be overwhelmed by information and influenced by emotional distress or cognitive bias during crises. The best solution is to combine AI’s data analysis and human judgment. AI helps, rather than replaces, human judgment in critical situations. Research suggests that neither humans nor AI alone can successfully manage complex disaster situations \cite{Visave2025}.
Trust, safety, and accountability are other vital issues in life-critical applications such as disaster response. Unless responders and the public trust the AI system’s output, the human actors will not take prompt action. Transparency is often a critical driver of trust for AI disaster platforms \cite{Visave2025}. Stakeholders must understand how an AI model is trained, where its training data came from, and its known capabilities. AI tools must be thoroughly vetted and validated across various disaster situations for safety. Competent reliability testing is necessary, as an unreliable AI for disaster applications can result in human casualties and a failure in response \cite{UMUT}. Accountability is also a significant issue: when human elements are involved in decision-making, there must be clarity on who is accountable for the outcome. The consensus is that such responsibility must lie with human operators or organizations rather than an algorithm. This is problematic in practice – where an AI recommendation leads to a negative outcome, the legal responsibility can be unclear. Initiatives such as the EU’s proposed AI Act and AI Liability Directive aim to bridge the gap by clarifying responsibilities for high-risk AI systems \cite{FeiFei}.

Disaster response is more dependent than ever on a cooperative connection between human experts and AI tools, especially on data-driven (DT)- enabled platforms, where the availability of real-time data poses challenges related to big data analytics \cite{geosciences8050165,rob}. 
Even with advanced simulations, human leadership, assisted by AI, remains the typical mode of operation in crises \cite{Elise, huang2023human, KatesMartuscello2025keeping}. This cooperation works best when the human-AI interface is well-established \cite{dc067a790b9c42108494facfcd75ffab}. AI can make recommendations at machine speeds, while human experts ensure that things are safe, ethical, and contextually correct. Disaster response can be more accurate by employing AI as a force multiplier rather than as a substitute for human decision-makers \cite{leybourn2025constraints,lmi2024hazard}.

\section{Discussion}

Implementing a disaster management system is not without serious difficulties. Despite recent developments, numerous technical, organizational, and ethical barriers remain. 
 The functioning of a digital city depends on high-quality and reliable data; yet, most cities are short of data or have poor data quality \cite{brucherseifer2025datataxonomyapplicabilitydigital}. Critical data for disaster risk (e.g., high-resolution elevation data, infrastructure databases) may be out-of-date, incomplete, or non-standard. In practice, some municipalities (mainly in developing countries) still lack digitlized data for essential assets like buildings or terrain. A lack of high-quality data can create blind spots within the DT and reduce the accuracy of simulations and warnings \cite{Gilman}.  Additionally, building a comprehensive data model for an entire city is a highly challenging task. There is no standard master plan yet – the terms and ideas are open to interpretation and ambiguous. Most recent studies recognize the application of DTs for disaster management as “still in its infancy,” with a lack of common understanding and high complexity due to the vast diversity of spatial and temporal scales \cite{info11120569}. It is technologically challenging to integrate different subsystems (transportation, utilities, communications) into a single interoperable platform, requiring customized integration and potentially leading to silos unless managed \cite{Tim,Rose}.

One of the challenges is related to governance and coordination. Disaster management system initiatives cut across many sectors (such as GDPR), and they can lead to jurisdictional and governance issues \cite{su16198337,Ford}. An effective disaster management system requires the collaboration of city governments, emergency services, utilities, and private sector partners. Therefore, establishing a disaster-resilient smart city necessitates coordinated policy and regulation across various domains. However, gaining funding and financial support from the government and other stakeholders is a significant challenge \cite{refId0,healthcare9020203}, in addition to issues related to data ownership, sharing agreements, and leadership accountability \cite{unknown,Socitm2025}.     
The increased interconnectivity and data exchange in Disaster management systems leads to privacy and cybersecurity issues. City-scale data platforms are potential targets for cyberattacks \cite{app13020790}. For example, examinations of European smart cities centered around a single IoT platform revealed that most of them did not meet basic cybersecurity standards. During a crisis, a data breach or cyber security attack would lead to a catastrophe(think false sensor readings or disabled communications) \cite{RomainSaccone2025}. Furthermore, pervasive sensors and surveillance (cameras, tracking apps) raise privacy concerns – large-scale data collection can lead to “surveillance capitalism” and corrupt citizens’ privacy and trust \cite{Curran,Mendoza03072022}, and the government can use data for suppressing revolutions and utilize data against the public. Balancing the condition for individuals’ data privacy in crises is an important issue \cite{smartcities6010027}.

There is a risk that advanced disaster management solutions can worsen social disparities if not implemented inclusively \cite{smartcities7060130}. Not everybody has access to technology or can benefit from digital services. Experts are warning that the benefits of DTs and intelligent technology can disproportionately accrue to some groups (or technology providers) at the expense of vulnerable groups. An example is highly automated systems displacing jobs or marginalizing those who are digitally untrained. It is crucial to ensure that tools like disaster apps or warning systems are accessible to the elderly, people with disabilities, or low-income groups (who may not have access to smartphones or up-to-date information). Understanding is another issue – explaining complex model outputs to the general public in simple language can be challenging \cite{smartcities8010023}. Without efforts to increase public digital literacy and involve residents in planning, even a technically sound system may not achieve its goals \cite{inbook}.

In practice, it is hard to gather and handle “big data.” \cite{Li2023}Most regions do not have the coverage or density of sensors to fill a DT in real-time completely; key data (e.g., underground infrastructure maps or real-time population densities) is not necessarily available at good quality resolution \cite{Lin2023}. Additionally, even if data is available, quality assurance (accuracy, timeliness, calibration) is complex to ensure; erroneous data would cause the DT to misrepresent reality \cite{EZKI}. A further limitation is the modeling complexity – disasters involve very complex physics and human behaviors, which are hard to model comprehensively \cite{Yu2022}. Modeling how every building in the city would respond to an earthquake or how thousands of people would behave during evacuation is a monumental task. Simplifications are inevitable; thus, the DT’s simulations will always have some error \cite{Grieves2017}.

 The computing infrastructure needed for a disaster management DT can be costly – both in the initial setup (sensors, data centers, specialized software) and ongoing operation (maintenance, data storage, continuous updates)and hiring Skilled personnel (data scientists, GIS experts, modelers)\cite{Jahangir}.In fast onset disasters (such as earthquakes or terrorist attacks \cite{Omnilert2025}), decisions must be made in seconds or minutes. Its utility is diminished if the DT’s computations \cite{ALNAZ} cannot keep up (due to latency or processing limits \cite{Jia}). Ensuring that DT systems remain robust and cyber-secure \cite{Haider,9566277} under extreme disaster conditions – where power outages or network failures may occur – is another practical limitation that designers must address.

\subsection{AI Agents for Real-Time Decision Making During Disasters}

AI agents are better than humans in processing big data streams and identifying patterns more rapidly, thus facilitating timely intervention. For instance, during the Turkish and Syrian earthquakes in Kahramanmaraş in 2023, AI systems processed satellite images to identify the most severely damaged zones in a short time, informing rescue units even before ground assessments were completed \cite{milvus2023ai_agents_disaster_management}. In the case of rapidly spreading wildfires, California's ALERT system, for example, utilizes an AI vision agent that surveys a network of more than 1,000 mounted cameras for subtle plumes of smoke in remote areas\cite{fuhrman2023california}. Such in-real-time AI surveillance is effectively an automatic lookout, often detecting fires earlier than the human eye. It is an example of how autonomous agents can provide new perspectives on the surroundings to facilitate timely disaster responses.

Beyond early detection, AI agents play a role in real-time decision optimization during unfolding emergencies. One method is through reinforcement learning (RL) agents that learn on the fly, adjusting response tactics dynamically . For instance, emergency routing agents, in the form of deep RL prototypes, reroute ambulances or fire units in real time, utilizing live traffic probes and road closure information\cite{milvus2023ai_agents_disaster_management}. By learning to bypass jammed roads or congested avenues, RL-based systems minimize response times in chaotic conditions. In the case of wildfires, simulations based on AI models predict fire growth in response to dynamic wind and fuel conditions, guiding incident commanders in prioritizing evacuations and distributing firefighting resources where they are most needed. Also, edge AI deployments can overcome latency and connectivity issues by processing information locally when connectivity is lost\cite{azfar2025enhancingdisasterresilienceuavassisted}.

\subsubsection{AI Agents for Coordinating Emergency Response Efforts}

Disaster response typically involves a diverse array of agents—firefighters, medical personnel, police, and volunteer groups—working in combination with autonomous agents, including drones, robots, and intelligent infrastructure. Coordinating heterogeneous assets in a dynamic situation is a complexity-laden coordination problem. AI agents, in the form of multi-agent systems (MAS), can be applied to coordinate communication, assignment, and collaboration among all tiers of the emergency response \cite{smythos2025utilizing}.

 Technical benefits of multi-agent coordination in disaster situations are as follows. Firstly, decentralized autonomous agents can conduct parallel actions. For example, several drone agents in a search-and-rescue operation may simultaneously scour various regions of a disaster area, looking for survivors or hazards \cite{kios2024glimpse}. Secondly, MAS architecture enables real-time information exchange and dynamic planning. The agents may pass important updates peer-to-peer (e.g., a drone locates a survivor's position and immediately informs nearby rescue team devices), enabling the overall system to reassign tasks dynamically. Third, multi-agency robustness is essential: even when single or several agents are compromised (due to destruction, loss of contact, etc.), other agents will still operate and re-adjust their approaches to bridge the gaps \cite{smythos2025utilizing}.

Another domain is evacuation management, where agent-based modeling (ABM) has been applied to simulate and improve evacuation plans \cite{smythos2025utilizing}. For example, in disaster situations, By running countless what-if scenarios, these agents help planners choose shelter locations that minimize travel distance for most people while avoiding overcrowding and allocate supplies (food, water, medical kits) across a region in a balanced way. Indeed, multi-agent optimization engines can treat resource allocation as a cooperative game, finding strategies that best meet the needs of all districts without requiring central planners to manually crunch data.

\subsubsection*{Utilizing VLM in IVSR}

 One crucial and effective technique in disaster management using deep learning (DT) is vision-language models (VLMs). These models may enhance wildfire management while profoundly improving situational awareness and decision support in fire emergencies.

The IVSR can ingest multi-source video/image feeds from indoor CCTV and outdoor cameras or drones. A vision-language analysis module (e.g., BLIP-2 or LLaVA) processes incoming frames to produce rich textual labels and descriptions of the scene \cite{Kaz}.
These labels include fire characteristics (intensity, spread rate), the type of burning objects (vehicle, furniture, vegetation), detected humans and their locations, and other spatial contexts. Advanced VLMs can generate structured scene descriptions that capture detailed context (e.g., environment type, fire state) in real-time. For example, BLIP-2 has demonstrated state-of-the-art performance in zero-shot image-to-text tasks \cite{li2023blip2bootstrappinglanguageimagepretraining}. LLaVA similarly connects a vision encoder to a large language model for general-purpose visual understanding \cite{liu2023visualinstructiontuning}. The system also leverages specialized detectors where needed (e.g., YOLOv7 for precise smoke/flame or human localization \cite{fire7040140}) to complement the VLM outputs.

Key steps for the data processing pipeline are as follows: Each fire incident video is processed frame by frame (or via keyframes) to extract semantic labels and metadata, forming an annotated scenario record. Modern VLMs (including GPT-4 Vision ) accept image inputs and can describe scenes with human-level detail, automatically labeling incident videos. The annotated scenarios are stored in a structured database (e.g., as JSON records or an ontology) indexed by features and embeddings. This growing repository serves as training data to continually refine the models. For instance, fine-tuning a 7B-parameter VLM on such structured fire data has improved accuracy in classifying fire scenes. The IVSR can periodically retrain or fine-tune its models on the accumulated data to improve its ability to recognize new fire situations. In real-time emergencies, the IVSR can perform a semantic similarity search against the scenario database to find past cases relevant to the current incident. This retrieval utilizes high-dimensional embeddings of the current scene (from the VLM or a CLIP-like model) and a vector index (e.g., FAISS \cite{chandra2024decisionsupportforestmanagement}) to efficiently fetch similar scenarios. The approach is analogous to case-based reasoning, as it assumes that "similar problems have similar solutions" \cite{Bannn}, as we explained in the introduction of IVSR.

While agentive AI holds the potential to transform DTs into proactive, self-governing command centers, some challenges remain: the need to make black box decision logic explainable and transparent, governance that provides attribution for autonomous actions, robust protection against manipulation, and a broad-based design to prevent biased results. Addressing these technical, regulative, and societal challenges frontally will be key to realizing the ultimate potential of AI-based, bidirectional systems for resilient and fair disaster response.

\section{Conclusion}

\label{sec:conclusion}

In this paper, we introduced a novel concept—the Intelligent Virtual Situation Room (IVSR)—as the core of a bidirectional DT approach to wildfire management. We validated IVSR’s core capabilities through detailed case‐study simulations, including real-time incident localization with dynamic rollback, privacy-preserving incident replay, collider-based fire-spread propagation, and on-site synthetic data generation for customized model retraining. 

Our IVSR concept builds upon this foundation by integrating a pre-computed library of disaster scenarios with AI‐driven similarity matching to recommend context‐specific intervention plans. This semiautomated decision-support capability accelerates response times and reduces operator cognitive load while preserving human oversight through an expert review panel. Collectively, these components turn conventional DT solutions into active, closed-loop systems capable of sending action-driven directions to both autonomous actuators and human teams.

In the future, we intend to scale up our indoor testbed to large-scale outdoor wildfire scenarios and add real-time actuation—in the form of autonomous drone deployment and dynamically reconfigurable sensor placement—to close the bidirectional feedback loop. We further intend to incorporate Agentic AI models for enhanced situational awareness and validate our method through field trials in collaboration with emergency organizations. Ultimately, our work paves the way for more intelligent and resilient disaster management systems that can rapidly and accurately forecast, simulate, and counteract natural disaster threats.

\bibliographystyle{model1-num-names}

\bibliography{cas-refs}

\appendix
\section{Literature Review Summary Tables}

This appendix contains tables summarizing the reviewed literature. Table \ref{tab:dt_fire_1} and \ref{tab:dt_fire_2} provide a comparison of recent Digital Twin approaches for fire management. Table \ref{tab:research_gaps} maps selected references to the research gaps they address.

\begin{table*}[htbp]
\centering
\footnotesize
\caption{Comparison of Recent(2024-2025) Digital-Twin Approaches for Fire Management}
\label{tab:dt_fire_1}
\rowcolors{2}{gray!10}{white}
\begin{tabularx}{\textwidth}{
    p{1.5cm}        
    Y             
    Y             
    Y             
    Y             
}
\toprule
\textbf{Reference} 
  & \textbf{Keywords} 
  & \textbf{Approach} 
  & \textbf{Key Contribution} 
  & \textbf{Case Study \& Notes}  \\
\midrule
Xie \textit{et al.} (2025) \citeyear{XIE2025103117}
  &  AIoT, Fire Forecast 
  & ADLSTM-Fire with LoRa sensors for building-wide mapping 
  & Enables real-time spatial visualization and forecasting of fire propagation 
  & FASA three-storey training facility \\

Zhang \textit{et al.} (2024) \citeyear{ZHANG2024100381}  
  & Tunnel fires, Deep learning, AIoT,  Fire safety management  
  & Transformer-based AIoT twin integrating sensor network,  Unity3D/WebGL  
  & Real-time prediction of fire location \& heat release rate
  & Full-scale SCFRI tunnel tests with n-heptane \& wood crib fires  \\

Zhou \textit{et al.} (2025)  \citeyear{Zhou_2025}
  & Remote sensing, Wildfire spread, Explainable AI  
  & Comparative CNN  vs. Transformer on 10 yr of U.S. satellite data + XAI 
  & Quantitative performance comparison; model interpretability; identified key drivers  
  & Decadal (2012–2020) 1 km resolution dataset for California \\

Huang \textit{et al.} (2024)  \citeyear{fire7110412}
  &  Wildfire detection, Simulation, VR visualization  
  & Systematic review and proposed Wildfire DT framework  
  & WFDT architecture; outlines R\&D challenges \& future directions  
  & conceptual model only, no field validation  \\

Shaposhnyk \textit{et al.} (2024) \citeyear{s24072366}
  & Next Generation First Responder;Cognitive Workload; Assistive Technologies
  & Extended SmartHub integrating on-body sensors, distributed ledger, and cloud-based AI
  & Embeds NGFR hub into smart-city infrastructure; designs personalized IPD assistant
  & Experimental sondage on CogLoad wristband dataset for CW monitoring \\

  Aydın \textit{et al.} (2025) \citeyear{Aydin2025}
  &  IoT; graph neural networks; LoRa; forecaster
  & GNN-based DT for LoRa-WSN throughput prediction, fed by an LSTM forecaster of network states
  & Decouples DT from two-way comms via forecaster; shows high accuracy in small-scale nets
  & Simulated clustered LoRa WSN  \\

  Li \textit{et al.} (2024) \citeyear{Li2024ReviewAP}
  &  wildland fire; emergency response; decision support; post-fire recovery; 
  & Systematic review of DT applications across wildfire prevention, monitoring, response, and recovery
  & Consolidates recent DT use‐cases; identifies key data sources (RS, IoT, AI, simulation)
  & Conceptual review only—no new empirical case  \\

Zhang \& Lee (2024) \citeyear{ZhangLee2024}
  &  Firefighting training; ML-based fire spread prediction; Real-time simulation; Escape route optimization
  &  3D fire‐scene DT driven by ML models for spread forecasting +  pathfinding for dynamic escape routing
  & Immersive, interactive DT for firefighter drills; real‐time fire‐and‐smoke updates
  & Three‐floor building scenario; trained on Algerian forest fires dataset  \\

Kim \textit{et al.} (2024) \citeyear{KIM2024105722}
  &   Dynamic time warping, Fire emergency management systems
  & Informative DT-based with cloud–edge architecture, pre-simulated fire-scenario knowledge
  & Comprehensive IDT architecture (attributes, behaviors, interfaces); online DTW-based fire recognition; zone-based risk assessment metrics 
  & Case study at Hoehyeon Underground Shopping Center, evacuation simulation under varying occupant loads \\

  Fan \textit{et al.} (2025) \citeyear{f16030519}
  &  moldering-to-flaming transition, Forest fire re-ignition, UAV, Fire suppression visualization
  & DT‐based StF‐trigger model with gradient‐descent‐fitted wind/slope parameters; dynamic 3D forest‐floor scenario
  & Predictive mapping of re‐ignition probability using historical data; UAV‐delivered hydrogel vs. solidified‐foam suppression
  & Virtual forests , 3D StF driving \& visualization\\

Zhang \textit{et al.} (2025) \citeyear{202520252025}
  & Differential Evolution, Deep Learning, Firefighting Force Configuration, Facility Layout
  & Iterative FSSM-based DT simulation integrating differential-evolution optimization for force allocation and DNN-guided facility layout
  & Proposes a joint optimization framework for pre-prevention forest rescue, modeling multi-resource interactions to reduce fire extinguishing time, burned area, and uncontrolled fire rate
  & Case studies on four forest maps (pure, mountain, basin, urban)  \\

Kamalakannan \textit{et al.} (2024) \citeyear{kamalaaa}
  &  Federated Learning, Decision Forest, Edge–Fog Computing, Fire Accident Mitigation 
  &  DT architecture with federated decision forest deployed across edge–fog nodes for real-time fire-accident detection 
  & Introduces a Federated Decision Forest for decentralized model training, plus an edge–fog–enhanced DT design with dynamic load-balanced resource allocation. 
  & 3D conveyor simulation with heat/dust sensors

\end{tabularx}
\end{table*}

\begin{table*}[htbp]
\centering
\footnotesize
\caption{Comparison of Recent(2024-2025) Digital-Twin Approaches for Fire Management}
\label{tab:dt_fire_2}
\rowcolors{2}{gray!10}{white}
\begin{tabularx}{\textwidth}{
    p{1.5cm}        
    Y             
    Y             
    Y             
    Y             
}
\toprule
\textbf{Reference} 
  & \textbf{Keywords} 
  & \textbf{Approach} 
  & \textbf{Key Contribution} 
  & \textbf{Case Study \& Notes}  \\
\midrule

Minasova \textit{et al.} (2024) \citeyear{Mina}
  & UAV, Aerial Photography, Forest Fire Monitoring
  & Mathematical model for correcting image distortions from AT UAV motion
  & Integrated system architecture and algorithm for building/updating a forest fire DT
  & Prototype evaluated on a synthetic UAV‐flight /fire‐spread dataset; software visualization only (no field trials) \\

Oktaj \textit{et al.} (2024) \citeyear{Fatma}
  & Forest fire detection, IoT networks, Graph Neural Network
  & LSTM‐based forecaster for per‐node sleep/transmit state prediction, 
  & Introduces a network-state forecaster to bridge IoT constraints and enable DT integration
  & Simulated clustered LoRa-like \& GPRS-like WSN over Akseki (Antalya), Turkey;\\

Noh \textit{et al.} (2025) \citeyear{Noh}
  & Network-controlled repeater, EV, LSTM, SVM
  & DT-based EV temperature monitoring using NCR-enabled wireless transfer + dual-spectrum CCTV
  & Early anomaly detection of EV thermal runaway
  & Pyeongchon underground parking lot (Anyang, Korea);simulated high-temp video via RTSP; prototype evaluation only \\

Wang \textit{et al.} (2024) \citeyear{RWFT}
  &  Remote sensing, Forest wildfire detection, Attention
  & RFWNet: DCNv3‐based backbone + dual‐path dynamic sparse attention + Vanilla Head
  & High‐fidelity synthetic\& real remote sensing wildfire datasets; 
  & Drone+edge (Jetson NX)+GPU deployment; detailed ablation \& real‐world trials \\

Qiang (2025) \citeyear{Asia}
  & Smart home, Fire monitoring, IoT  
  & Wireless sensor array (temp \& smoke) → Zigbee/Wi-Fi → DT model of home → decision‐support engine  
  & First home-scale DT fire‐safety framework; real-time fire‐spread simulation, false-alarm reduction, sensor-layout optimization  
  &  kitchen/hall/bedroom/living-room deployment  \\

\end{tabularx}
\end{table*}

\begin{table*}[htbp]
\centering
\caption{Research Gaps Addressed by Selected References (2023--2025) in DT-Based Disaster Management. \textbf{RTDI} = Real-Time Data Integration; \textbf{BC} = Bidirectional Communication/Feedback; \textbf{HS} = High-Fidelity Simulation \& Scenario Modeling; \textbf{AI/ML} = Artificial Intelligence/Machine Learning-based Prediction \& Anomaly Detection; \textbf{DMS} = Data Management, Scalability \& Interoperability; \textbf{CP} = Cybersecurity \& Privacy Concerns; \textbf{PI} = Practical Implementation/Case Study. A check mark (\(\checkmark\)) indicates that the paper addresses the respective research gap.}
\label{tab:research_gaps}
\small
\begin{tabular}{l c c c c c c c c}
\toprule
\textbf{Reference} & \textbf{Year} & \textbf{RTDI} & \textbf{BC} & \textbf{HS} & \textbf{AI/ML} & \textbf{DMS} & \textbf{CP} & \textbf{PI} \\
\midrule
Riaz et al. \citeyear{s23052659}                    & 2023 & $\checkmark$ & $\checkmark$ & $\checkmark$ & $\checkmark$ &  &  &  \\ \
Ariyachandra \& Wedawatta \citeyear{sss}            & 2023 & $\checkmark$ & $\checkmark$ & $\checkmark$ & $\checkmark$ & $\checkmark$ & $\checkmark$ &  \\
Zhong et al. \citeyear{nhess-23-1755-2023}           & 2023 & $\checkmark$ &  & $\checkmark$ & $\checkmark$ & $\checkmark$ &  & $\checkmark$ \\
Salinas et al. \citeyear{10155819}                  & 2023 & $\checkmark$ & $\checkmark$ & $\checkmark$ & $\checkmark$ &  &  & $\checkmark$ \\
El-Shafeiy et al. \citeyear{ElShafeiy2023}          & 2023 & $\checkmark$ &  &  & $\checkmark$ &  &  & $\checkmark$ \\
Khajavi et al. \citeyear{10130166}                  & 2023 & $\checkmark$ & $\checkmark$ &  &  &  & $\checkmark$ &  \\
Cheng et al. \citeyear{buildings13051143}           & 2023 & $\checkmark$ & $\checkmark$ & $\checkmark$ & $\checkmark$ & $\checkmark$ & $\checkmark$ &  \\
Lawal \citeyear{Lawal}                              & 2023 & $\checkmark$ &  &  &  & $\checkmark$ & $\checkmark$ & $\checkmark$ \\
Li et al. \citeyear{Li2023}                         & 2023 & $\checkmark$ &  & $\checkmark$ & $\checkmark$ & $\checkmark$ &  &  \\
Lin et al. \citeyear{Lin2023}                       & 2023 & $\checkmark$ &  &  &  & $\checkmark$ &  &  \\
Othman et al. \citeyear{Othman}                     & 2023 & $\checkmark$ & $\checkmark$ & $\checkmark$ &  & $\checkmark$ & $\checkmark$ & $\checkmark$ \\
Yassin et al. \citeyear{YASSIN2023100039}           & 2023 & $\checkmark$ & $\checkmark$ & $\checkmark$ &  & $\checkmark$ & $\checkmark$ &  \\
Accarino et al. \citeyear{Accarino_2023}            & 2023 &  &  &  & $\checkmark$ &  &  &  \\
\midrule
Huang et al. \citeyear{fii}                         & 2024 & $\checkmark$ &  & $\checkmark$ & $\checkmark$ &  &  &  \\
Li et al. \citeyear{gii}                            & 2024 & $\checkmark$ &  & $\checkmark$ & $\checkmark$ & $\checkmark$ &  &  \\
Shadrin et al. \citeyear{Shadrin2024}               & 2024 & $\checkmark$ &  & $\checkmark$ & $\checkmark$ &  &  &  \\

Rychlicki et al. \citeyear{app14209243}             & 2024 & $\checkmark$ &  &  &  &  &  &  \\
Xu et al. \citeyear{xu2024leveraginggenerativeaiurban} & 2024 &  &  & $\checkmark$ & $\checkmark$ &  &  &  \\
Shariatpour et al. \citeyear{doi:10.1061/JUPDDM.UPENG-4650} & 2024 &  &  & $\checkmark$ &  &  &  &  \\
Adreani et al. \citeyear{Adr}                       & 2024 & $\checkmark$ & $\checkmark$ &  &  & $\checkmark$ &  &  \\
An et al. \citeyear{suan16219482}                   & 2024 & $\checkmark$ &  &  & $\checkmark$ &  &  & $\checkmark$ \\
McCourt et al. \citeyear{court2024usedigitaltwinssupport} & 2024 &  &  &  & $\checkmark$ &  &  &  \\
Noeikham et al. \citeyear{Piya}                     & 2024 & $\checkmark$ &  &  &  & $\checkmark$ &  &  \\
Han \& Choi \citeyear{han2024designing}             & 2024 &  &  & $\checkmark$ &  &  &  &  \\
Abraham et al. \citeyear{abraham2024evacuationmanagementframeworksmart} & 2024 & $\checkmark$ &  & $\checkmark$ &  &  &  & $\checkmark$ \\
Shahbazi \& Bunker \citeyear{SHAHBAZI2024102780}    & 2024 & $\checkmark$ &  &  &  &  & $\checkmark$ &  \\
Kolotouchkina et al. \citeyear{smartcities7060130}   & 2024 &  &  &  &  & $\checkmark$ &  &  \\
Jahangir et al. \citeyear{Jahangir}                  & 2024 &  &  &  &  & $\checkmark$ &  &  \\
Menges et al. \citeyear{menges2024predictivedigitaltwincondition} & 2024 & $\checkmark$ &  &  & $\checkmark$ &  &  &  \\
Alnaser et al. \citeyear{app142412056}              & 2024 & $\checkmark$ &  &  & $\checkmark$ & $\checkmark$ &  &  \\
Arowoiya et al. \citeyear{AROWOIYA2024641}          & 2024 &  &  & $\checkmark$ &  & $\checkmark$ &  &  \\
Dihan et al. \citeyear{DIHAN2024e26503}             & 2024 &  &  &  &  & $\checkmark$ &  &  \\
Robinson et al. \citeyear{robinson2024twinetconnectingrealworld} & 2024 &  & $\checkmark$ &  &  &  &  &  \\
Yang et al. \citeyear{Yang2024}                     & 2024 & $\checkmark$ &  & $\checkmark$ &  &  &  &  \\
El-Agamy et al. \citeyear{El-Agamy2024}             & 2024 &  &  &  &  & $\checkmark$ &  &  \\
Mir \citeyear{mir}                                  & 2024 & $\checkmark$ & $\checkmark$ & $\checkmark$ & $\checkmark$ & $\checkmark$ & $\checkmark$ &  \\
Mylonas et al. \citeyear{mylonas2024facilitatingaioperatorsynergy} & 2024 &  & $\checkmark$ &  & $\checkmark$ &  &  &  \\
LAGAP et al. \citeyear{LAGAP2024104629}             & 2024 &  &  & $\checkmark$ &  & $\checkmark$ &  & $\checkmark$ \\
Chia et al. \citeyear{ijgi13120445}                 & 2024 & $\checkmark$ &  & $\checkmark$ &  &  &  &  \\
Shaposhnyk et al. \citeyear{s24072366}              & 2024 & $\checkmark$ &  &  &  &  &  &  \\
Kim et al. \citeyear{KIM2024105722}                 & 2024 & $\checkmark$ &  & $\checkmark$ & $\checkmark$ &  &  & $\checkmark$ \\
\midrule 
Cheruku et al. \citeyear{CHeraku}                   & 2025 & $\checkmark$ &  &  &  & $\checkmark$ &  & $\checkmark$ \\
Socitm \citeyear{Socitm2025}                        & 2025 &  &  &  &  & $\checkmark$ &  &  \\
Romain \& Saccone \citeyear{RomainSaccone2025}      & 2025 &  &  &  &  &  & $\checkmark$ &  \\
Raes et al. \citeyear{inbook}                       & 2025 &  &  & $\checkmark$ &  & $\checkmark$ &  & $\checkmark$ \\
Electronics Review \citeyear{electronics14040646}   & 2025 & $\checkmark$ & $\checkmark$ & $\checkmark$ & $\checkmark$ & $\checkmark$ &  &  \\
Yang et al. \citeyear{yang2025leveraginglargelanguagemodels} & 2025 &  &  &  & $\checkmark$ & $\checkmark$ &  &  \\
Qanazi et al. \citeyear{smartcities8010023}         & 2025 &  &  &  &  & $\checkmark$ &  &  \\
Ouedraogo et al. \citeyear{EZKI}                   & 2025 &  &  &  &  & $\checkmark$ &  &  \\
Setijadi Prihatmanto et al. \citeyear{Seti}         & 2025 & $\checkmark$ &  & $\checkmark$ &  &  &  & $\checkmark$ \\
Fan et al. \citeyear{f16030519}                     & 2025 & $\checkmark$ &  & $\checkmark$ & $\checkmark$ &  &  & $\checkmark$ \\
Aydin \& Oktug \citeyear{Aydin2025}                 & 2025 & $\checkmark$ &  &  &  & $\checkmark$ &  & $\checkmark$ \\
\bottomrule
\end{tabular}
\end{table*}

\section*{Data Availability}
The data from the case company that support the findings of this study are available from the corresponding author upon reasonable request.

\end{document}